\documentclass{article}
\usepackage{spconf,amsmath,graphicx}

\usepackage{float}
\usepackage{enumitem}
\setlist{nosep, leftmargin=14pt}

\usepackage{mwe} 

\title{High Spectral Spatial Resolution Synthetic HyperSpectral Dataset form multi-source fusion}
\name{Yajie Sun$^1$, Ali Zia$^{2,3}$, Jun Zhou$^1$\thanks{The multimodal dataset is available at https://github.com/spectral-3D-lab/multimodal-dataset.git.  }}
\address{$^1$Griffith University\
$^2$CSIRO, Australia\
$^3$The Australian National University}
\begin{document}
%
\maketitle



This research paper introduces a synthetic hyperspectral dataset that combines high spectral and spatial resolution imaging to achieve a comprehensive, accurate, and detailed representation of observed scenes or objects. Obtaining such desirable qualities is challenging when relying on a single camera. The proposed dataset addresses this limitation by leveraging three modalities: RGB, push-broom visible hyperspectral camera, and snapshot infrared hyperspectral camera, each offering distinct spatial and spectral resolutions. Different camera systems exhibit varying photometric properties, resulting in a trade-off between spatial and spectral resolution. RGB cameras typically offer high spatial resolution but limited spectral resolution, while hyperspectral cameras possess high spectral resolution at the expense of spatial resolution. Moreover, hyperspectral cameras themselves employ different capturing techniques and spectral ranges, further complicating the acquisition of comprehensive data. By integrating the photometric properties of these modalities, a single synthetic hyperspectral image can be generated, facilitating the exploration of broader spectral-spatial relationships for improved analysis, monitoring, and decision-making across various fields. This paper emphasizes the importance of multi-modal fusion in producing a high-quality synthetic hyperspectral dataset with consistent spectral intervals between bands.

\section{Multimodal Dataset}

\begin{figure}[htp]

\begin{minipage}[b]{1.0\linewidth}
  \centering
  \centerline{\includegraphics[width=6.5cm]{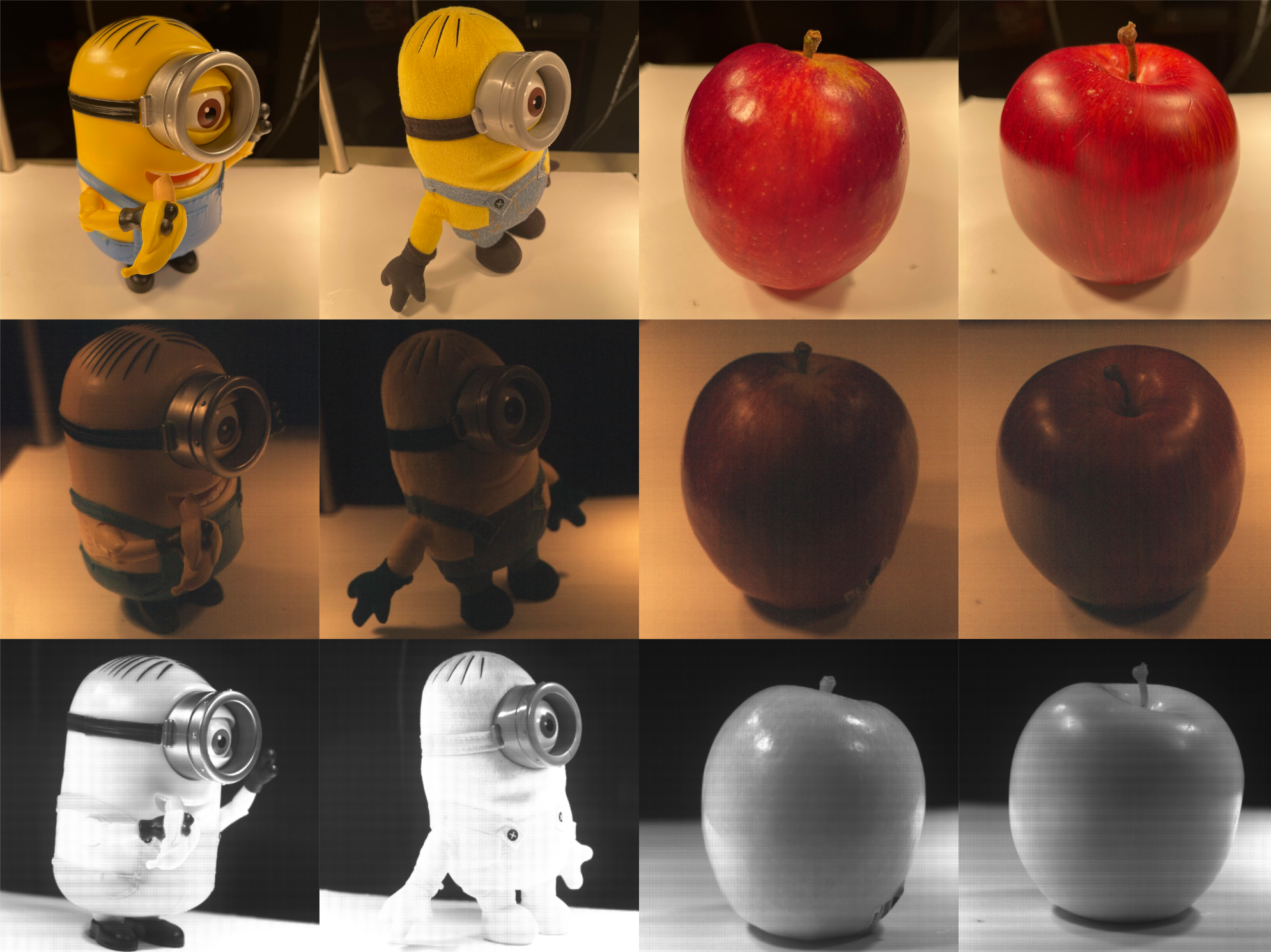}}
\end{minipage}
\caption{A plastic toy and a plush toy. A plastic apple and a real apple. The first, second, and third columns are RGB, hyperspectral and infrared images respectively.}
\label{fig:res}
\end{figure}
The image synthesis process encompasses the careful consideration of various factors, including generating new spectral bands and up-scaling spatial resolution. To ensure consistency in spatial resolution, a spatial resolution adjustment is applied to up-sample low-resolution hyperspectral images, aligning them with the high-resolution RGB images. Moreover, to have a consistent spectral gap, we synthesize additional spectral bands, by the extraction of relevant spectral features from neighboring bands that aid in inferring missing spectral information~\cite{2023busifusion}. 

The push-broom visible hyperspectral camera has 3650*2048 spatial resolution and 470-900nm spectral range with 150 spectral bands. The snapshot infrared camera has 2048*1088 spatial resolution and 600-975nm spectral range with 25 spectral bands. The acquisition of RGB images is done by an iPhone 12pro with 1200 megapixels. Twelve objects of different materials were photographed from 60 different angles. The materials include plastic, resin, rubber, tin, paper, and real fruit. Examples are shown in Fig. 1.

This dataset can help researchers investigate the spectral features of different materials before purchasing specialized cameras. Furthermore, these images are captured from different views and can be used to generate complete 3D hyperspectral models by Structure from Motion\cite{zia20153d}. Therefore, the potential applications of multi-source synthesis data are extensive. 


\bibliographystyle{IEEEbib}
\bibliography{strings,refs}

\end{document}